# Semantic-Based VPS for Smartphone Localization in Challenging Urban Environments

Max Jwo Lem Lee, Li-Ta Hsu, Hoi-Fung Ng, Shang Lee

*Abstract*— **Accurate smartphone-based outdoor localization system in deep urban canyons are increasingly needed for various IoT applications such as augmented reality, intelligent transportation, etc. The recently developed feature-based visual positioning system (VPS) by Google detects edges from smartphone images to match with pre-surveyed edges in their map database. As smart cities develop, the building information modeling (BIM) becomes widely available, which provides an opportunity for a new semantic-based VPS. This article proposes a novel 3D city model and semantic-based VPS for accurate and robust pose estimation in urban canyons where global navigation satellite system (GNSS) tends to fail. In the offline stage, a material segmented city model is used to generate segmented images. In the online stage, an image is taken with a smartphone camera that provides textual information about the surrounding environment. The approach utilizes computer vision algorithms to rectify and hand segment between the different types of material identified in the smartphone image. A semantic-based VPS method is then proposed to match the segmented generated images with the segmented smartphone image. Each generated image holds a pose that contains the latitude, longitude, altitude, yaw, pitch, and roll. The candidate with the maximum likelihood is regarded as the precise pose of the user. The positioning results achieves 2.0m level accuracy in common high rise along street, 5.5m in foliage dense environment and 15.7m in alleyway. A 45% positioning improvement to current state-of-the-art method. The estimation of yaw achieves 2.3° level accuracy, 8 times the improvement to smartphone IMU.**

*Index Terms*—**Localization, Navigation, Smartphone, VPS, Urban Canyons, Pedestrian, GNSS, BIM, 3D Building Models**

## 1. Introduction

U RBAN localization is an essential step to the development of numerous IoT applications such as digital management of navigation, augmented reality, commercial related services [1], and an indispensable part of our daily lives due to its widespread application [2]. For indoor areas, Wi-Fi based localization has become extremely popular and many researchers are focused in this area [3-5]. However, the use of Wi-Fi in urban areas is still very challenging, suffering tens of meters even in strong signal conditions [6]. As indicated in [7], the calibration of Wi-Fi fingerprinting database and the density of Wi-Fi beacons in urban areas pose a lot of challenges. As a result, Wi-Fi is mostly suitable for indoor positioning. In the context of outdoor pedestrian localization, the application of global navigation satellite system (GNSS) is the key technology to provide accurate positioning/timing service in open field environments. Unfortunately, its positioning performance in urban areas still has a lot of potential to improve due to signal blockages and reflections caused by tall buildings and dense foliage [8]. In such environments, most signals are non-line-of-sight (NLOS) which can severely degrade the localization accuracy [9]. Hence, they cause large estimation errors if they are either treated as line-of-sight (LOS) or not used properly [10]. Therefore, efforts have been devoted to developing accurate urban positioning systems in recent years. A review on state-of-the-art localization is published in 2018 [11]. Each of these technologies has its advantages and limitations. However, some of these solutions face other challenges, such as mobility, accuracy, cost, and portability. For a pedestrian self-localization system, it should be accurate and efficient enough to provide positioning information [12]. Nowadays, a personal smartphone is equipped with various embedded sensors, such as gyroscope, accelerometer, vision sensors, etc. These sensors can be used for urban localization. The requirement of being inexpensive, easy to deploy, and user friendly are also satisfied.

With the rise of smart cities, 3D city models have been developing rapidly and have become widely available [13]. An idea called GNSS shadow matching was proposed to improve urban positioning [14]. It first classifies the received satellite visibility by the received signal strength and then scans the predicted satellite visibility in the vicinity of the ground truth position. Position is then estimated by matching the satellite visibilities. Another method uses ray-tracing–based 3DMA GNSS algorithms that cooperate with pseudorange have been proposed [15]. The integration of the shadow matching and range-based 3DMA GNSS are proposed in [16]. Where the performance of this approach in multipath mitigation and NLOS exclusion depends on the accuracy of the 3D building models [17].

This work is supported by PolyU RISUD on the project – BBWK "Resilient Urban PNT Infrastructure to Support Safety of UAV Remote Sensing in Urban Region."

M.J.L. Lee, L-T. Hsu and H-F. Ng are with Interdisciplinary Division of Aeronautical and Aviation Engineering, The Hong Kong Polytechnic University (PolyU). L-T. Hsu is also with Research Institute for Sustainable Urban Development (RISUD). S. Lee is with Hong Kong University of Science and Technology School of Engineering. Corresponding author: Li-Ta Hsu (e-mail: lt.hsu@polyu.edu.hk).

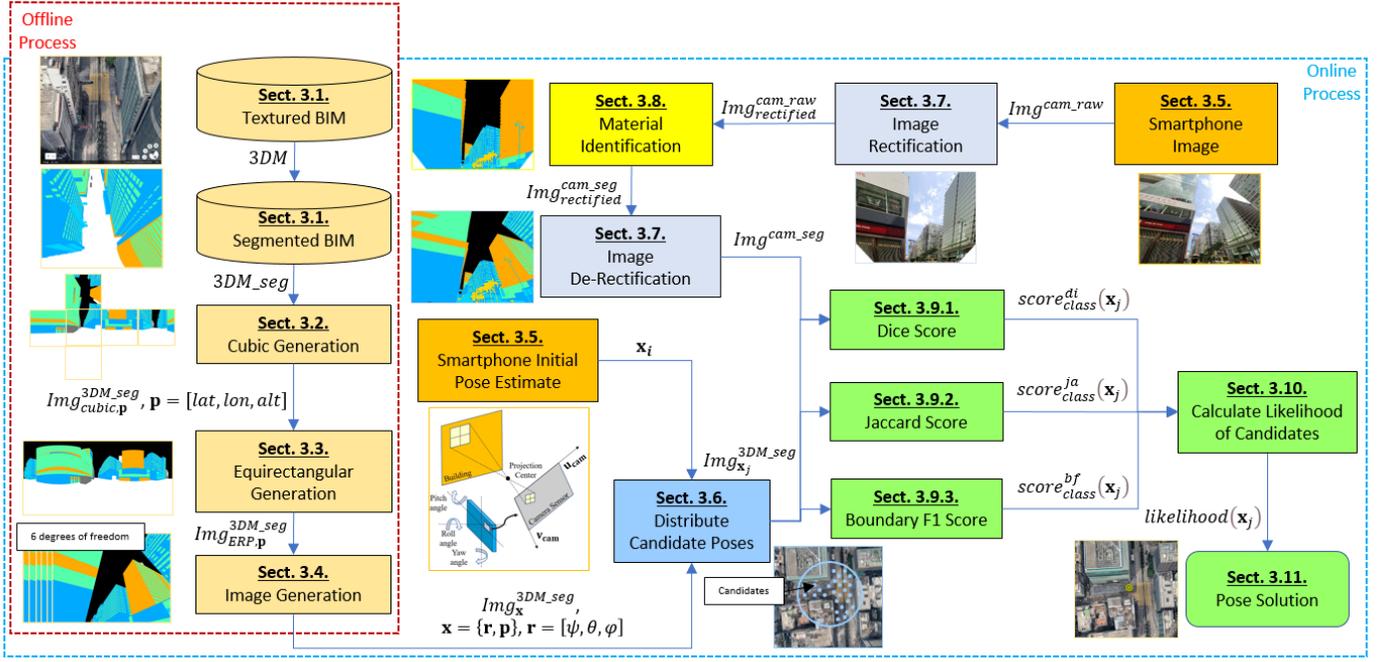

Fig. 1. Flowchart of the proposed semantic-based VPS based on segmented smartphone image and segmented generated images.

In the last few years, there has been an increasing interest in inferring position using 3DMA and vision-integrated methods. The motivation is that these are complementary methods, which in combination can provide rich scenery information. The major reason is that high-performance modern smartphones provide cameras, and computing platform for storage, data processing, and fusion, which can be easily exploited. The general idea behind most of these approaches is to find the closest image to a given query picture in a database of pose-tagged images (three-dimensional position and three-dimensional rotation, adding up to six degrees of freedom [DOF]).

Research have demonstrated that it is possible to obtain precise positioning by matching between a camera image and a database of images. One popular approach uses a sky-pointing fisheye camera equipment to detect obstacles and buildings in the local environment [18]. When it is used in conjunction with image processing algorithms, they allow the matching of building boundary skyplot (skymask) to obtain a position and heading.

Currently, there are several studies that make use of smartphone image to estimate the pose of the user. Google's recently developed feature-based visual positioning system (VPS) identifies edges within the smartphone image and matches with edges captured from pre-surveyed images in their map database [19]. The pose-tagged edges are stored in a searchable index, and require updates overtime by the users. Another area of study focuses on semantic information, such as identifying static location-tagged objects (doors, tables, etc.) in smartphone images for indoor positioning [20], however reference objects are often limited in the outdoor environments. Thus, other researchers have studied the use of skyline or building boundaries to match with smartphone images [21-24]. This provides a mean positional error of 4.5m and rotational error of 2-5° at feature rich environments [21].

Although both methods are suitable in urban areas where GNSS are often blocked by high rise, the former uses features extracted from pre-surveyed images for precise localization, suffers from image quality dependency, and requires frequent updates using the cloud-sourcing data from users. While the latter suffers from obscured or non-distinctive skyline, prominent in highly urbanized areas where dynamic objects dominate the environment. Thus, detection solely on the edges and skyline may not be enough for practical use and precise positioning. Standing at the point of view of how a pedestrian navigate him/herself, in addition to the identification of features and skyline, we, human beings, also locate based on the visual landmarks that consists of different semantic information, and each semantic has a material of its own.

Inspired from both existing methods, our proposed novel solution is the semantic-based VPS by utilizing different types of materials that are widely seen and continuously distributed in urban scenes. The proposed method offers several major advantages over the existing methods.

- Firstly, we can take advantage of building materials as visual aids for precise self-localization, overcoming inaccuracies due to non-distinctive or obscured skyline, which are common in urban environments.

- Secondly, with the use of building information modelling (BIM), it does not require pre-surveyed images, hence it is highly scalable and low cost.

- Thirdly, unlike storing feature data as point clouds in a searchable index, the semantics of materials are stored as a vector map, making it simple to update and label accurately.

- Lastly, the proposed method identifies and considers dynamic objects into its scoring system, which are usually neglected in previous studies.

Thus, the paper is an interdisciplinary research paper that integrates the knowledge of BIM, geodesy, image processing, and navigation. We believe this interdisciplinary research demonstrates a very good solution to provide seamless positioning for many futuristic IoT applications.

The remainders of the paper are organized as follows. Sect. 2 explains the overview of the proposed semantic-based VPS approach. Sect. 3 describes the candidate image generation, material identification and image matching in detail. Sect. 4 describes the experimentation process and the improvement of the proposed algorithm verified with existing advanced positioning methods. Sect. 5 contains the concluding remarks and future work.

## 2. Overview of the Proposed Method

An overview of the proposed semantic-based VPS method is shown in Fig. 1. The method is divided into two main stages: an offline process, and an online process.

In the offline process, the building models are manually segmented into different colors based on the material, which grantees a perfect representation of the materials in the 3D city model (Sect. 3.1). The segmented city model is used to generate cubic projection at each position (Sect. 3.2), they are then converted into the equirectangular projection (Sect. 3.3), to generate an image at each pose (Sect. 3.4). By storing the images in an offline database within the smartphone, we can derive a memory-effective representation of accurate reference images suitable for smartphone-based data storage.

Based on the generated images, we propose a semantic-based VPS method for smartphone-based urban localization. In the online process, the user captures an image with their smartphone (Sect. 3.5), with the initial pose estimated by the smartphone GNSS receiver, and IMU sensors (Sect. 3.5). Then, candidates (hypothesized poses) are spread across a searching grid based on the initial pose (Sect. 3.6). The smartphone image is then rectified (Sect. 3.7), and segmented based on the identified types of materials (Sect. 3.8). The segmented smartphone image is de-rectified (Sect. 3.7) and matched with the candidate images using multiple metrics to calculate the similarity scores (Sect. 3.9). The scores of each method are combined to calculate the likelihood of each candidate. (Sect. 3.10). The chosen pose is determined by the candidate with the maximum likelihood among all the candidates (Sect. 3.11). The details of the proposed method are described in the following section.

## 3. Proposed Method in Detail

### 3.1. Textured & Segmented BIM

The city model used in this research is provided by the *Surveying and Mapping Office, Lands Department, Hong Kong* [25]. It consists of only buildings and infrastructures; foliage and dynamic objects are not represented in the models. Each building model consists of level of detail (LOD) 1-3, stored as Virtual Reality Modeling Language (VRML) format. Each building model has its own corresponding 2D vector map in JPG format that provides textural information of the building.

The building vector maps were manually labelled, in which each pixel in the texture image is assigned a color for the material it represents, which can then be used to simulate a segmented 3D city model as shown in Fig. 2. In this research, we used six classes in total to test the feasibility of the proposed method, each class has their own respective RGB color: Sky (black), Stone (blue), Glass (green), Metal (orange), Foliage (yellow), Others (light blue). The building vector maps were labelled manually with the *Image Labeler* application which is part of the *Computer Vision Toolbox, MATLAB* [26].

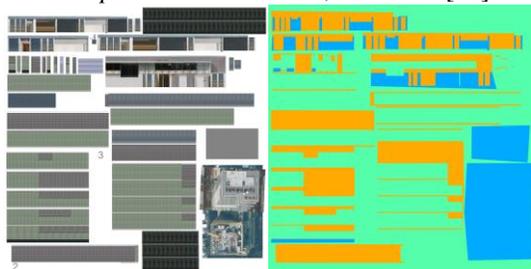

Fig. 2. Example of building vector map on the left and the corresponding segmented building vector map on the right.

This process will only be conducted once, until the building is modified physically in the real environment. It should be noted that the vector maps only contain four material classes, Stone, Glass, Metal and Others. Sky is identified by the empty space in the 3D city model simulation. Foliage is only recognized in the smartphone image in the online process.

The city model uses the 3D Cartesian meter coordinate system on a plane to determine the positioning coordinates. Therefore, it was necessary to convert the measured GNSS positioning information in (latitude and longitude) back to the 3D Cartesian coordinate. Thus, we transform between the WGS84 Geographic coordinates and Hong Kong 1980 Grid coordinates using the equations described by the *Surveying and Mapping Office, Lands Department, Hong Kong* [27].

### 3.2. Cubic projection generation

Each projection and its respective coordinate systems require careful clarification. Cubic projection is a method of environment mapping that utilizes the six faces of a cube in a 3D Cartesian coordinate system. The environment is projected onto the sides of a cube and stored as six squares. The cube map is generated by first rendering the scene of a position six times each from a viewpoint, with the views defined by a 90-degree angle of view frustum representing each cube face shown in Fig. 1.

Six 90° view frustum square images were captured within *Simulink 3D Animation, MATLAB* [28] with a virtual camera at each defined position to map a cubic projection. The defined positions store the latitude, longitude, and altitude. Equation (1) denotes the generation process.

$$\mathbf{p} = [lat, lon, alt]$$
$$Img^{3DM\_seg}_{cubic,\mathbf{p}} = C\_P(3DM\_seg, \mathbf{p}) \qquad (1)$$

Where $\mathbf{p}$ is the three-dimensional position, $3DM\_seg$ is the segmented building model, and $C\_P$ is the function to capture the six images. The cubic projection at a defined position is denoted as $Img^{3DM\_seg}_{cubic,\mathbf{p}}$.

### 3.3. Equirectangular projection generation

The utilization of equirectangular projection is to assist the faster generation of images in the offline stage. Equirectangular projection is a common sphere-to-plane mapping, which also allows a full spherical view of its surrounding shown in Fig. 1.

Equation (2) shows the transforming between cubic and equirectangular projection at given position, which requires the conversion from Cartesian coordinates to spherical coordinates.

$$Img_{ERP,\mathbf{p}}^{3DM\_seg} = ER\_P\left(Img_{cubic,\mathbf{p}}^{3DM\_seg}\right) \quad (2)$$

Where $ER\_P$ is the function to convert cubic projection into equirectangular projection described in [29]. The equirectangular projection at a defined position is denoted as $Img_{ERP,\mathbf{p}}^{3DM\_seg}$.

Just like the cubic projection, the defined equirectangular projection positions store the latitude, longitude, and altitude.

For practicality, the relationships between the Cartesian coordinates and spherical coordinates can be precomputed. This means that the angles' information can be retrieved by mapping given Cartesian coordinates in the precompute lookup table, reducing offline computational load and time.

### 3.4. Image generation

To match with the smartphone image coordinate system, further transformation from equirectangular projection to image (also called gnomonic projection) is required. The image coordinate system uses the 2D Cartesian system [ $\mathbf{u}, \mathbf{v}$ ] described in (3) and (4).

$$\mathbf{r} = [\psi, \theta, \varphi]$$
$$\mathbf{x} = \{\mathbf{p}, \mathbf{r}\}$$
$$Img_{\mathbf{x}}^{3DM\_seg} = RL\_P\left(Img_{ERP,\mathbf{p}}^{3DM\_seg}, \mathbf{r}\right) \quad (3)$$

Where $\mathbf{r}$ is the three-dimension rotation, $\mathbf{x}$ is the state that defines the pose which holds the position and rotation. $RL\_P$ is the function to convert equirectangular projection into image projection described in [30]. The image at a defined pose is denoted as $Img_{\mathbf{x}}^{3DM\_seg}$. The format of the images can be described as:

$$Img_{\mathbf{x}}^{3DM_{seg}} = SI(\mathbf{u_x}, \mathbf{v_x})$$
$$SI \in \left\{ \begin{matrix} \text{Sky (0), Stone (1), Glass (2),} \\ \text{Metal (3), Foliage (4), Others (5)} \end{matrix} \right\} \quad (4)$$

Where $\mathbf{u_x}, \mathbf{v_x}$ are the 2D pixel coordinates of the pixel inside the image generated based on the pose $\mathbf{x}$. $SI$ is the function that assigns each pixel an indexed number to represent a material class. Each image stores its corresponding pose. Fig. 1 shows an example of an image generated from the equirectangular projection based on a defined pose. It is important to mention that the cubic and equirectangular projections are only used to assist image generation. The generated images are then stored in the smartphone as indexed images to reduce storage size and used in the online phase for image matching.

To match the pose precisely, the generated images must have the same intrinsic parameters as the smartphone image described in Sect. 3.5.

### 3.5. Smartphone image acquisition and format

Since the smartphone image is analyzed according to the urban scene, the comparison is likely to perform well when there is a richer and more diverse urban scene. Therefore, the widest available angle lens is the preferred choice as it is more suitable to capture more information of the surrounding urban scene in the image. A conventional smartphone camera with a 120° diagonal field of view, 4:3 aspect ratio, resolution of [1000,750] pixels is used to capture images shown in Fig. 1.

### 3.6. Candidate Pose Distribution

Candidate poses are distributed around the initial estimated pose. The initial rough estimation of the pose is calculated by the smartphone GNSS receiver and IMU when capturing an image with the smartphone. The candidate latitudes and longitudes are distributed around the initial position in a 40m radius with 1m resolution. The candidate altitudes remain the same as measured by the smartphone due to its already high accuracy. The candidate rotation is distributed around initial rotation with 30° yaw, 3° pitch and 3° roll with 1° separation. The following distribution values were calibrated by finding the maximum possible error when comparing the smartphone estimated rotation with their ground truth. The poses will then be reduced to the specific candidate poses shown in (5).

$$\mathbf{X} = \{\mathbf{x_0} \cdots \mathbf{x_s}\} \quad (5)$$

Where $s$ is the index of the poses outside of the buildings, that is generated offline and saved in a database. Candidate pose $\mathbf{x}_j$ is extracted from the database $\mathbf{X}$, where $\mathbf{x}_j \in \mathbf{X}$, and the subscript $j$ is the index of the candidate poses. The corresponding image for each candidate pose is denoted as $Img_{\mathbf{x}_j}^{3DM\_seg}$. The distributed candidate images will then be used to compare against the smartphone image.

### 3.7. Image Rectification/De-Rectification

Since the feasibility of recognizing objects' appearances greatly benefits from the normalization [31], therefore the initial camera rotation information is used to perform rectification on the images as a preparation for further material recognition. The smartphone images are transformed into the rectified form. The rectification approximates the smartphone image to a unified view that ideally is horizontal and vertically level. Fig. 3 shows how an image is captured with the smartphone.

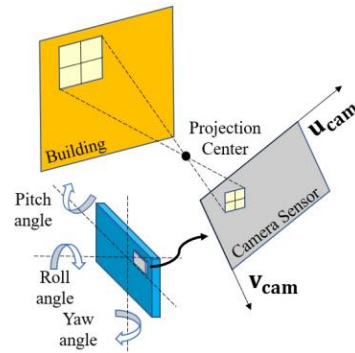

Fig. 3. Demonstration of an image captured with a pitch and roll

angle.

The proposed image rectification assumes that the rotation of the camera image is approximately known from the output of the smartphone IMU. From this, horizon and keystone correction can be performed. The first kind of distortion is associated to the roll angle of the camera, whereas the second kind is due to the camera pitch angle. The roll and pitch angle, associated respectively to horizon and keystone distortion, can be corrected, whereas the yaw angle cannot.

Horizontal correction is obtained by first rotating the image around its centre point by the opposite of the initial estimated roll angle. Afterwards, keystone correction is achieved by geometrical transform denoted in (6).

$$Img_{rectified}^{cam\_raw} = rectification(Img^{cam\_raw}) \quad (6)$$

Where $rectification$ is the function to transform the smartphone image $Img^{cam\_raw}$ with horizontal and keystone correction into the rectified image described in [31]. The rectified image is denoted as $Img_{rectified}^{cam\_raw}$.

The greater the object elements that are further away from the horizon is, the greater the distortion is. However, the horizon area of the rectified images, which usually contains more distinctive features, provides a more suitable input for classification. Once combined, it can rectify the image such that it is an approximation image taken at horizontal and vertical level shown in Fig. 3. Then after segmentation, the image can be de-rectified with the reverse of the image rectification process.

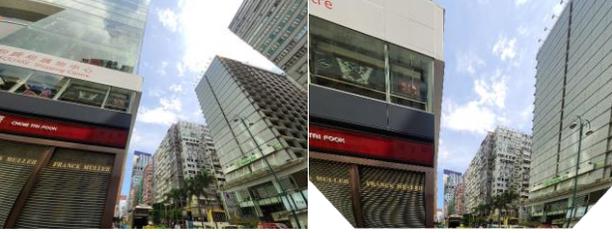

Fig. 4. Example of smartphone image captured with 45° pitch on the left, rectified image on the right.

It should be noted, horizontal and keystone correction respectively requires the knowledge of the camera intrinsic parameter (principal point, focal lengths).

### 3.8. Hand Labelled Material Segmentation

The captured smartphone images were labelled manually with the Image Labeler application in MATLAB. In the future, however, we plan to utilize a deep learning neural network to identify the material automatically. This will be discussed in further detail in Section 5. The rectified smartphone image will then be hand labelled to output the ideally segmented smartphone image illustrated in (7).

$$Img_{rectified}^{cam_{seg}} = H\_L(Img_{rectified}^{cam_{raw}}) \quad (7)$$

Where $H\_L$ is the function to segment the rectified image manually. The segmented rectified image is denoted as $Img_{rectified}^{cam_{seg}}$. After segmentation, the image can be de-rectified with the reverse of the image rectification equations described in (8).

$$Img^{cam\_seg} = De\_Rectification(Img_{rectified}^{cam\_seg})$$
$$Img^{cam\_seg} = SI(\mathbf{u_{cam}}, \mathbf{v_{cam}}) \quad (8)$$

Where $De\_Rectification$ geometrically transforms the segmented rectified image to normal segmented image $Img^{cam\_seg}$, shown in Fig. 1. $\mathbf{u_{cam}}, \mathbf{v_{cam}}$ are the 2D pixel coordinates of the pixel inside the image captured based on the smartphone camera.

### 3.9. Material Matching

In the online stage, the candidate images are compared to the smartphone image. The matching algorithm calculates the score of each candidate image. The target function is to find the candidate image with the largest similarity with respect to the semantic information of materials. A usual approach is to use the region and contours of each material class in the candidate image to compare with the corresponding material class in the smartphone image.

#### 3.9.1. Dice Metric

We used Sørensen–Dice coefficient metric to compare the region of two material segmented images [32]. Equation (9) shows the calculation of the similarity index for each material class.

$$sim_{class}^{di}\left(Img^{cam_{seg}}, Img_j^{3DM seg}\right)$$
$$= \frac{\left|Img_j^{3DMseg}(class) \cap Img^{cam seg}(class)\right|}{0.5\left(N_{class,j}^{3DM seg} + N_{class}^{camera_{seg}}\right)} \quad (9)$$

$$class \ni \{sky, stone, glass, metal, foliage, others\}$$

Where $class$ is the index that represents a material, and $sim_{class}^{di}\left(Img^{cam_{seg}}, Img_j^{3DM seg}\right)$ is the similarity index of the smartphone image and the candidate image for a material class.

A measure to consider is the ratio of the detected region compared to the total image size. A smaller matched region should have lower weighting, whereas a larger matched region should have higher weighting. Therefore, the similarity for each segmented material needs to be weighted according to the number of pixels they occupy in the candidate image to calculate the score of each class represented in (10).

$$N_{class,j}^{3DMseg} = \left|Img_j^{3DMseg}(class)\right|$$
$$score_{class}^{di}(\mathbf{x}_j) = sim_{class}^{di}(Img^{cam_{seg}}, Img_j^{3DM\_seg})$$
$$\cdot \left(N_{class,j}^{3DMseg} / N_{total}\right) \quad (10)$$

Where $N_{class,j}^{3DMseg}$ is the pixel region of a material class in the candidate image, and $N_{total}$ is the total number of pixels in an image. The dice score of a class is denoted as $score_{class}^{di}(\mathbf{x}_j)$.

Finally, the score for each material is combined to become the score of the candidate shown in (11).

$$score^{di}(\mathbf{x}_j) = \sum_{class} score_{class}^{di}(\mathbf{x}_j) \quad (11)$$

#### 3.9.2. Jaccard Metric

The Jaccard coefficient metric is similar to the Dice coefficient metric, but satisfy the triangle inequality and measures the intersection over the union of the labelled region instead [33]. We used the Jaccard coefficient metric to also compare the region of two material segmented images. Equation (12) demonstrates the calculation of the similarity index for each material class.

$$sim_{class}^{ja}\left(Img^{cam\_seg}, Img_j^{3DM_{seg}}\right)$$
$$= \frac{\left|Img_j^{3DM_{seg}}(class) \cap Img^{cam_{seg}}(class)\right|}{\left|Img_j^{3DM_{seg}}(class) \cup Img^{cam_{seg}}(class)\right|} \quad (12)$$

Where $sim_{class}^{ja}\left(Img^{cam_{seg}}, Img_j^{3DM_{seg}}\right)$ is the similarity index of the smartphone image and the candidate image for a material class. Just like the former metric, the similarity for each segmented material needs to be weighted according to the number of pixels they occupy in the candidate image to calculate the score of each class, represented in (13).

$$score_{class}^{ja}(\mathbf{x}_j) = sim_{class}^{ja}(Img^{cam\_seg}, Img^{3DM\_seg})$$
$$\cdot \left(N_{class,j}^{3DM_{seg}} / N_{total}\right) \quad (13)$$

The score of a class is denoted as $score_{class}^{ja}(\mathbf{x}_j)$. Finally, the score for each material is combined to become the score for each candidate shown in (14).

$$score^{ja}(\mathbf{x}_j) = \sum_{class} score_{class}^{ja}(\mathbf{x}_j) \quad (14)$$

### 3.9.3. Boundary F1 Metric

The contour quality significantly contributes to the perceived segmentation quality, the Boundary F1 (BF) metric benefits in that it evaluates the accuracy of the segmentation boundaries [34]. Which are not captured by the Dice and Jaccard metrics, as they are regional-based metrics.

Let us call $B^{cam_{seg}}(class)$ the boundary of the class of $Img^{cam_{seg}}(class)$, and likewise $B_j^{3DM_{seg}}(class)$ the boundary of the class of $Img_j^{3DM_{seg}}$. For a distance threshold of 5 pixels, the metric disregards the content of the segmentation beyond the threshold distance of 5 pixels under which boundaries are matched. The precision for a class is defined as:

$$P_{class}(\mathbf{x}_j)$$
$$= \frac{1}{\left|B_j^{3DM_{seg}}\right|} \sum_{b \in B_j^{3DM_{seg}}(class)} \llbracket d\left(b, B^{cam_{seg}}(class)\right)$$
$$< 10 \rrbracket \quad (15)$$

The recall for a class is defined as:

$$R_{class}(\mathbf{x}_j)$$
$$= \frac{1}{|B^{cam_{seg}}|} \sum_{b \in B^{cam_{seg}}(class)} \llbracket d\left(b, B_j^{3DM_{seg}}(class)\right)$$
$$< 10 \rrbracket \quad (16)$$

With $\llbracket \ \rrbracket$ as the Iversons bracket notation, where $\llbracket s \rrbracket = 1$ if $\llbracket s \rrbracket = true$ and 0 otherwise, and $d()$ denotes the Euclidean distance measured in pixels. The Boundary F1 measure for a class is given by:

$$score_{class}^{bf}(\mathbf{x}_j) = \frac{2 \cdot P_{class}(\mathbf{x}_j) \cdot R_{class}(\mathbf{x}_j)}{R_{class}(\mathbf{x}_j) + P_{class}(\mathbf{x}_j)} \quad (17)$$

The BF score of a class is denoted as $score_{class}^{bf}(\mathbf{x}_j)$. Finally,

the score for each material is combined by averaging the score over all classes present in the candidate image to become the total score for each candidate shown in (18).

$$score^{bf}(\mathbf{x}_j) = \frac{1}{n\_class} \sum_{class} score_{class}^{bf}(\mathbf{x}_j) \quad (18)$$

Where $n\_class$ is the total number of classes, in this research, we used six classes.

### 3.10. Combined Material Matching

We considered the score of each method (Dice, Jaccard, BF) for the 9 tested images described in Sect. 4 to calibrate their respective CDF based on a Gaussian distribution. The scores of each method is used to calculate the corresponding probability value in their respective distributions.

$$prob^*(\mathbf{x}_j) = \frac{1}{\sigma^* \cdot \sqrt{2\pi}} \cdot \int_{-\infty}^{score^*(\mathbf{x}_j)} e^{-\frac{1}{2}\left(\frac{x-\mu^*}{\sigma^*}\right)^2} dx \quad (19)$$

TABLE I . Parameters of Gaussian distribution

| Method | Standard Deviation | Mean |
|--------|--------------------|------|
| Dice | 0.1813 | 0.6686 |
| Jaccard | 0.1567 | 0.5399 |
| BF | 0.1387 | 0.4275 |

Where $*$ is the variable that is dependent on the method, $\sigma$ is the standard deviation and $\mu$ is the mean of the CDF.

The combined probability becomes the likelihood of each candidate.

$$likelihood(\mathbf{x}_j) = prob^{di}(\mathbf{x}_j) \cdot prob^{ja}(\mathbf{x}_j)$$
$$\cdot prob^{bf}(\mathbf{x}_j) \quad (20)$$

### 3.11. Pose Solution

A higher priority is given to the candidate image with a higher likelihood. In theory, the candidate image at ground truth should have the maximum likelihood. Thus, the candidate with the maximum likelihood is selected as the chosen candidate indicated in (21).

$$\hat{\mathbf{x}} = \arg\max_{\mathbf{x}_j}\left(likelihood(\mathbf{x}_j)\right) \quad (21)$$

Where $\arg\max_{\mathbf{x}_j}$ is a function that filters the highest total score, and $\hat{\mathbf{x}}$ is the estimated candidate pose with the highest likelihood. The chosen candidate pose stores the latitude, longitude, altitude, yaw, pitch, and roll.

TABLE Ⅱ. Locations and images tested with the proposed semantic-based VPS method

| Loc. | Experimental Images | | | |
|---|---|---|---|---|
| 1 | The Hong Kong Polytechnic University, Hung Hom | | | |
| | Overview | 1.1 | 1.2 | 1.3 |
| | 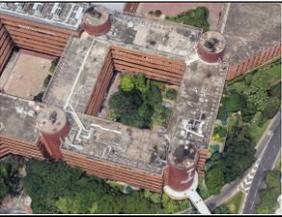 | 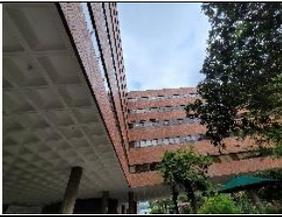 | 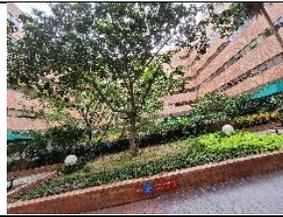 | 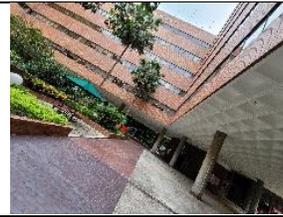 |
| | Overview | Obscured | Concealed | Obscured |
| 2 | Isquare, Tsim Sha Tsui | | | |
| | Overview | 2.1 | 2.2 | 2.3 |
| | 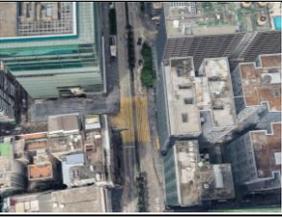 | 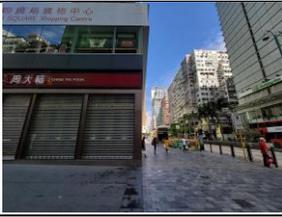 | 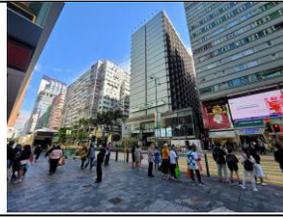 | 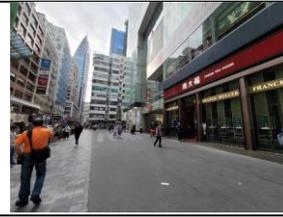 |
| | Overview | Distinctive | Distinctive | Distinctive |
| 3 | East Tsim Sha Tsui | | | |
| | Overview | 3.1 | 3.2 | 3.3 |
| | 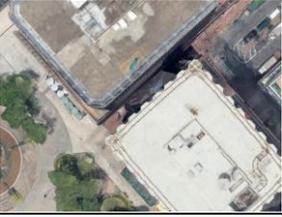 | 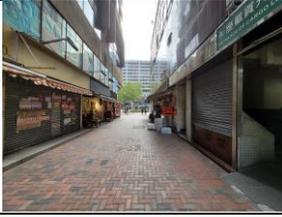 | 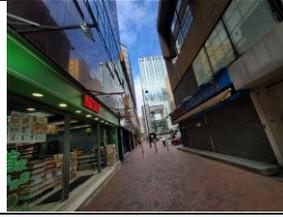 | 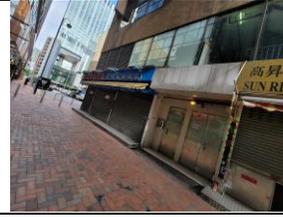 |
| | Overview | Symmetrical | Insufficient | Insufficient |

## 4. EXPERIMENTAL RESULTS

### 4.1. Image and test location setting

In this study, the experimental locations were selected within the Tsim Sha Tsui and Hung Hom area of Hong Kong, as shown in Table Ⅱ. Three locations were selected in challenging deep urban canyons surrounded by tall buildings where GNSS signals are heavily reflected and blocked. Three images were taken at each of the selected locations using a generic smartphone camera (Samsung Galaxy Note 20 Ultra 5G smartphone with the ultra-wide-lens 13mm 12-MP f/2.2) and a tripod. The experimental ground truth positions were determined based on *Google Earth* and nearby identifiable landmarks, such as a labelled corner on the ground. Based on the experience of previous researches [18, 35], the ground truth uncertainty of latitude and longitude is $\pm 1m$ and yaw is $\pm 2°$. The pitch and roll angles are measured using the *XPRO geared head, Manfrotto*, with $\pm 1°$ uncertainty, respectively.

The experimental images were chosen with the following skyline categorizations: distinctive, symmetrical, insufficient, obscured and concealed. Categorizations were based on the difficulties experienced by current 3DMA GNSS and vision-based positioning methods. The smartphone was used to capture the images and to record the low-cost GNSS position and IMU rotation. The GNSS receiver within the smartphone was a Broadcom BCM47755. The IMU was a LSM6DSO MEMS and was designed by STMicroelectronics. Images were taken at each location with different combinations of scenic features to demonstrate the proposed semantic-based VPS method. The locations were chosen to test the following environments respectively, dense foliage (Loc. 1), along street (Loc. 2), and alleyway (Loc. 3).

### 4.2. Positioning results using ideal segmentation

The positioning quality of the proposed method was analyzed based on the ideal smartphone image segmentation. The experimental results were then post-processed and compared to the ground truth and different positioning algorithms, including:

1. Proposed semantic-based VPS (Combination of Dice, Jaccard and BF Metrics)
2. Proposed semantic-based VPS (Dice only)
3. Proposed semantic-based VPS (Jaccard only)
4. Proposed semantic-based VPS (BF only)
5. Skyline Matching: Matching using sky and building class only [21].
6. 3DMA: Integrated solution by 3DMA GNSS algorithm on shadow matching, skymask 3DMA and likelihood based ranging GNSS [36].
7. WLS: Weighted Least Squares [37].

8. NMEA: Low-cost GNSS solution by Galaxy S20 Ultra, Broadcom BCM47755.

TABLE Ⅲ. Positioning performance comparison of the proposed semantic-based VPS and other advanced positioning algorithms.

| Loc. | Deviation from Ground Truth Error. Unit: meter. | | | | |
|------|------|------|------|------|------|
| | Semantic-based VPS (Combined) | Skyline Matching | 3DMA | WLS | NMEA |
| 1.1 | 7.07 | 22.92 | 7.96 | 17.66 | 36.24 |
| 1.2 | 4.34 | 22.62 | | | |
| 1.3 | 5.28 | 7.14 | | | |
| **1. Avg.** | **5.56** | **17.56** | | | |
| 2.1 | 0.66 | 14.80 | 6.87 | 23.29 | 7.94 |
| 2.2 | 1.83 | 1.58 | | | |
| 2.3 | 3.43 | 2.89 | | | |
| **2. Avg.** | **1.97** | **6.42** | | | |
| 3.1 | 29.89 | 13.57 | 18.80 | 46.58 | 18.89 |
| 3.2 | 6.61 | 25.53 | | | |
| 3.3 | 10.53 | 24.80 | | | |
| **3. Avg.** | **15.68** | **21.30** | | | |
| **All Avg.** | **7.74** | **15.09** | **11.21** | **29.18** | **21.02** |

**Loc. 1** is in an urban environment with dense foliage, which contains multiple non-distinctive medium rise buildings. The results show the positioning accuracy of the proposed semantic-based VPS improves upon the existing advanced positioning methods. An error of approximately 5.56 meters from the smartphone ground truth suggests that the semantic-based VPS can be used as a positioning method in foliage dense environments. Utilizing additional material information from buildings, it outperforms skyline matching thrice as much. The inability of skyline matching was due to the presence of foliage obscuring the skyline. Without an exposed skyline, it cannot match correctly and risks increasing the positioning error. 3DMA has shown to correct the positioning to a higher degree, coming behind the proposed method. The positioning error of WLS and NMEA were likely because of the diffraction of GNSS signals passing under the foliage with the combination of high-rise buildings.

Shown in the heatmap in Table Ⅴ, the proposed method using Dice and Jaccard have very large positioning errors, possibly due to the lack of distinctive materials captured in the smartphone image. The tested location is surrounded by buildings of the same shape, size, and material. Therefore, it is a very challenging environment for the proposed method as the candidate images share a common material distribution. It can be seen in this situation, using the BF achieves a higher positioning accuracy over the Dice and Jaccard, as it calculates the material contour rather than the material region. Thus, with the combination of the three metrics, this foliage dense environment proved suitable for the proposed method, which successfully utilize materials as information for matching.

Loc. 2 is in a common along street urban environment with high rise buildings. The results show the positioning accuracy of the proposed method improves the positioning accuracy to around two-meter level accuracy. In an environment where skyline matching should perform the best, the proposed method outperforms skyline matching over thrice as much as well. The matching of diverse material distributed in the scene in addition to the distinctive skyline has significantly improved positioning accuracy. 3DMA lacks behind skyline matching slightly, while WLS has increased the positioning error. It should be noted that the estimated positioning error for the NMEA is around 8m, significantly smaller than that of Loc. 1. This is likely due to the relative open area along the street as shown in Table Ⅱ.

The heatmap results shown in Table Ⅴ has demonstrated that the metrics complement each other when combined. As shown in Loc. 2.1, in a scene with diverse materials, the Dice and Jaccard have a higher positioning accuracy and achieve a higher likelihood over BF. Therefore, the combination of the three metrics leans towards the regional based similarities.

Loc. 3 is by far the most challenging urban environment for the 3DMA GNSS and vision-based positioning methods due to the closely compact high rise and visually symmetrical features. It can be seen all methods suffer in this environment, and most noticeably WLS. The results show that the positioning error of the proposed method is nearly 16 meters and can be improved significantly. It should be noted that this is still a 35% positioning improvement compared to skyline matching. Due to the lack of a distinctive skyline, skyline matching can potentially risk increasing the positioning error if matched with the wrong image, demonstrated at this position. 3DMA lacks behind the proposed method, and as demonstrated, only the proposed method and 3DMA improves the positioning accuracy slightly.

The poor results can be explained by two conditions required for accurate positioning. Firstly, the images ideally should have no segmentation error. This error is not considered in the positioning results, as we are assessing the ideal image segmentation. Instead, we have analyzed the segmentation error in relation to the positioning error is Sect. 4.4. Secondly, ideally there should be no discrepancies between the smartphone image and the candidate image at ground truth. Loc. 3 suffers from the latter as shown in Table Ⅳ.

TABLE Ⅳ. Discrepancy between reality and 3D city model

| | Reality | 3D City Model |
|------|---------|---------------|
| Textured | 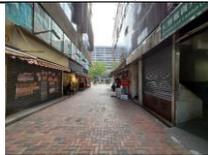 | 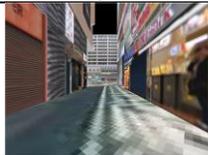 |
| Labelled | 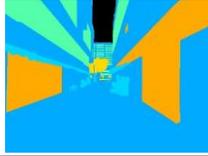 | 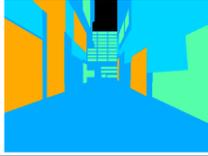 |

This error is shown in the positioning results of **Loc. 3**, where many candidates share a common similarity and color. Thus, it is important to ensure the 3D city is constantly updated to reflect reality.

TABLE Ⅴ. Heatmap on the likelihood of candidate images compared to the smartphone image based on the proposed semantic-based VPS method

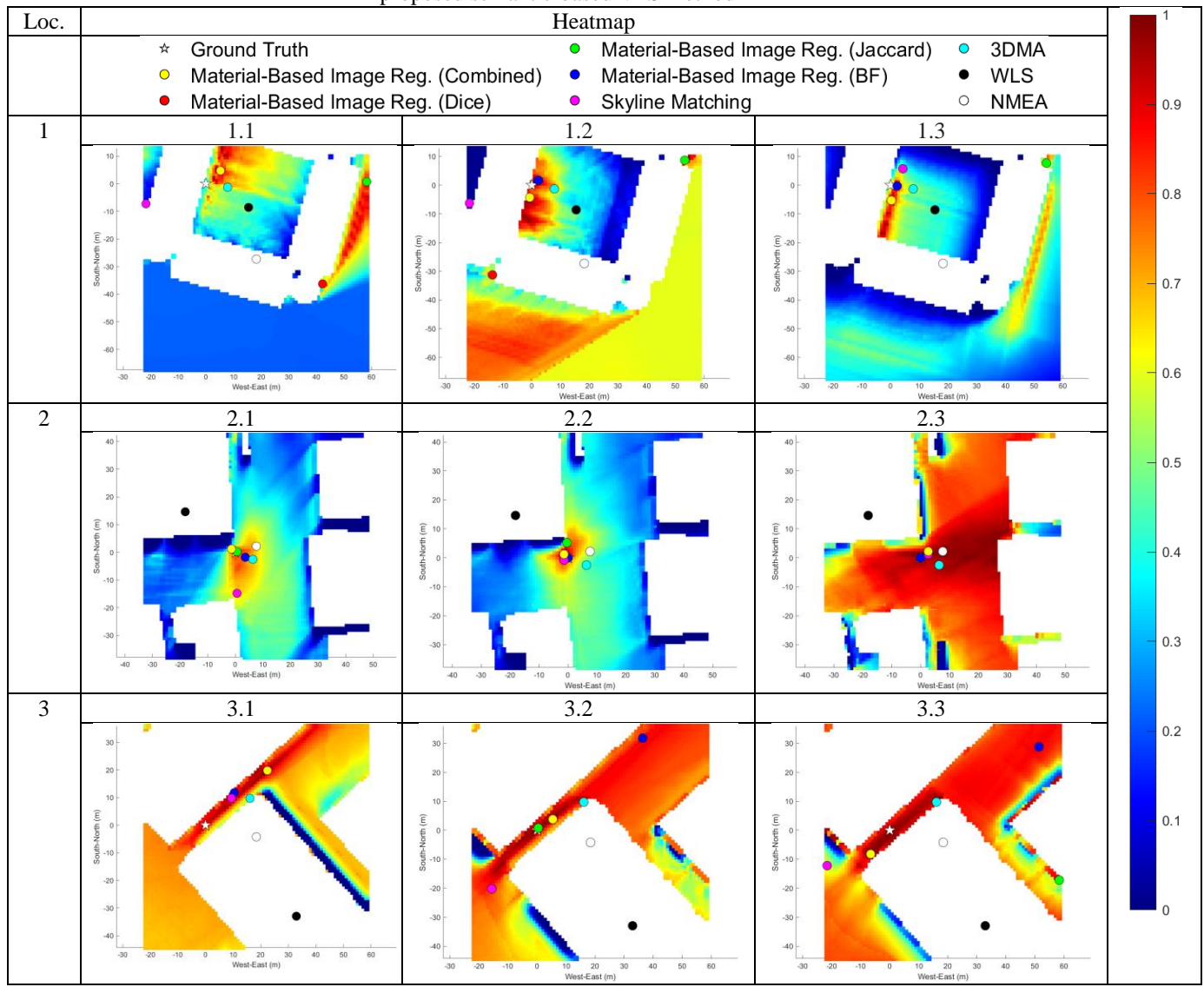

## 4.3. Rotational results using ideal segmentation

The three-dimensional rotational performance of the proposed method was analyzed based on the ideal smartphone image segmentation, then compared to the smartphone IMU.

TABLE Ⅵ. Rotational performance comparison of the proposed semantic-based VPS and smartphone IMU

| Loc. | Deviation from Ground Truth Error. Unit: degrees. | | | | | |
|---|---|---|---|---|---|---|
| | Semantic-based VPS | | | Smartphone IMU | | |
| | $\psi$ | $\theta$ | $\varphi$ | $\psi$ | $\theta$ | $\varphi$ |
| 1.1 | -4 | 0 | -1 | -27 | -2.0 | 1.0 |
| 1.2 | 3 | 2 | -2 | 7 | 0.5 | -0.5 |
| 1.3 | 3 | 2 | -1 | 18 | -0.5 | 0.5 |
| **1. Avg.** | **3.3** | **1.3** | **1.3** | 17.3 | 1.0 | 0.6 |
| 2.1 | 5 | 1 | -2 | 11 | 0.5 | -1.0 |
| 2.2 | -3 | -1 | 0 | 18 | 2.0 | 0.0 |
| 2.3 | 1 | 2 | -2 | 19 | -2.0 | 0.5 |
| **2. Avg.** | **3** | **1.3** | **1.3** | 16 | 1.5 | 0.5 |
| 3.1 | 2 | 2 | -2 | 31 | 1.0 | -1.5 |
| 3.2 | 0 | 1 | 0 | 28 | 0.5 | -0.2 |
| 3.3 | 0 | -2 | -2 | 27 | -0.5 | -0.2 |
| **3. Avg.** | **0.6** | **1.7** | **1.3** | 28.6 | 0.6 | 1.8 |
| **All Avg.** | **2.3** | **1.4** | **1.3** | **20.6** | **1.0** | **1.0** |

The results show that, in an urban environment with features, the material of buildings can be used to estimate the rotation. The yaw, pitch and roll have an accuracy of 2.3, 1.4 and 1.3 degrees, respectively. However, the smartphone IMU pitch and roll estimation is already very accurate compared to the proposed method, and thus the proposed method will only degrade the estimation. Instead, the proposed method succeeds at predicting the yaw accurately within an average of 2.3 degrees. Hence, the proposed method can be considered an accurate approach to estimate the heading of the user in an urban environment.

Therefore, it is suggested that the proposed method should use the already accurate altitude, pitch and roll for position and yaw estimation. Eliminating the estimation of three dimensions will significantly reduce computational load as less candidate images are used for matching.

### 4.4. Segmentation accuracy vs localization results

To test out how the semantic segmentation accuracy affects the localization results, we considered the two conditions required for accurate positioning. Ideally, there should be no segmentation error and no discrepancies between the smartphone image and the candidate image at ground truth. We can therefore further classify these two types of errors: contour-based error and regional-based error. We have tested in our experiments that discrepancies can contribute heavily to the positioning accuracy, shown in Table IV where the smartphone image differs from the candidate image at ground truth. Therefore, we can consider this as a regional-based error as the entire region differs between the images. We should also consider the contour-based error, which is not demonstrated in our experiments, but is reflected in a realistic output of a semantic segmentation neural network where the boundaries of a region are shifted. Contour error can be problematic for boundary related metrics such as the BF metric, which focus on the evaluation along the object edges. Getting these correct is very important, any shift in alignment can lead to mismatch with another candidate image. Thus, we considered the candidate images at ground truth to be the ideal images, as there is no regional-based error nor contour-based error. We purposely mislabeled the ideal images by adding the two types of noise to model the amount of segmentation accuracy.

To model the two types of errors, we performed a Monte-Carlo simulation. We elastically distorted the ideal image randomly to generate over 1000 distorted images described in [38], each with a distinctive regional-based and contour-based error. We then compared the distorted image with the ideal image using two metrics, the combined Dice and Jaccard metric for regional-based error, and the BF metric for the contour-based error. We then used our proposed method to obtain a positioning error by comparing the positioning solution of the distorted image with the ground truth position. Fig. 5 shows the candidate image with contour mislabeled using the elastic distortion algorithm.

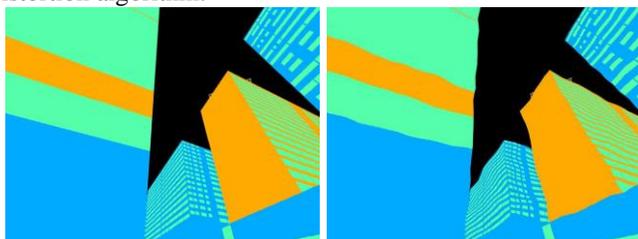

Fig. 5. Example of a candidate image on the left, and a slightly elastically distorted candidate image on the right.

**Characteristics of Positioning Error vs Segmentation Error**

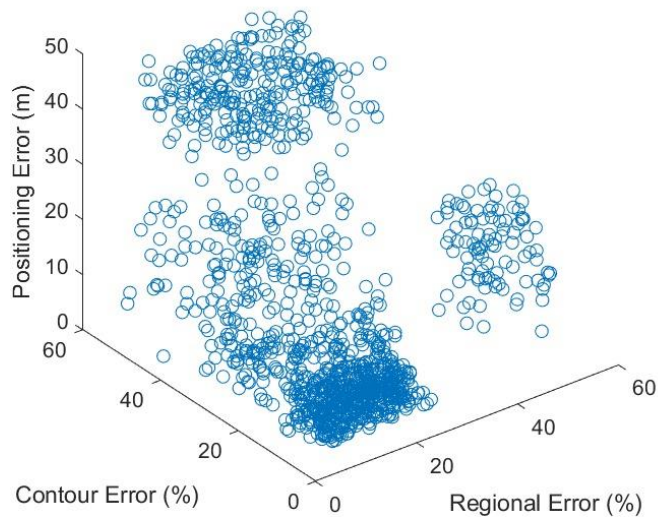

Fig. 6. The effects of contour-based error and regional-based error on the positioning error of the proposed semantic-based VPS.

The results show a good positioning accuracy at lower levels of segmentation error. It can be seen the positioning error in the 0 to 20% segmentation error range is approximately 0-5 meters. However, the proposed method begins to suffer when incorrect segmentation reaches more than 20% for contour-based error and 25% for regional-based error. This is followed by a deteriorating positioning performance, where the positioning error spreads to 10-20 meters. At 40% contour and regional-based error, the matching algorithm fails to perform accurately, and risk increasing the positioning error. It can be seen at this segmentation error range, the distorted image matches with random incorrect candidate images, thus the positioning error spreads across a wide region.

The Monte-Carlo simulation results demonstrate the importance of a correct contour-based and regional-based segmentation, and suggests that to successfully utilize the proposed method with a high positioning accuracy, a semantic segmentation neural network with no less than 80% segmentation accuracy is preferred. The results also suggest disabling the proposed method when the smartphone image is matched with a candidate image with a segmentation difference of more than 20-25%. In such situations, relying on other advanced positioning techniques such as 3DMA would likely yield better positioning results.

### 4.5. Discussion on Validity and Limitation

The proposed method presented in this research permits to self-localize based on material that is widely distributed among the urban scenes. Provided that the smartphone image segmentation is ideal, experiments show that our approach outperform the performance of positioning by 45% compared to current state of the art methods and improves the performance of yaw by 8 times compared to smartphone IMU sensors.

The pitch and roll estimated by the proposed method, however, achieves a lower performance by half a degree compared to the smartphone IMU sensors. Hence, it is suggested that the proposed method use the already accurate pitch and roll estimated by the smartphone IMU sensors. The elimination of altitude, pitch and yaw estimation will significantly reduce computational load as less candidate images are used for matching.

Another limitation comes from inaccurate segmentation. As demonstrated in this research, the 3D model was out of date, leading to discrepancies between the smartphone image and candidate image at ground truth. It has been shown when the segmentation error is greater than 20-25%, the positioning performance deteriorates significantly. Therefore, it is necessary to update the utilized 3D city model frequently.

## 5. Conclusions and Future Work

### 5.1. Conclusions

This paper proposes a novel semantic-based VPS solution for pose (six-DOF) estimation by introducing materials as a new source of information. In short, the semantic information of materials is extracted from the smartphone image and compared to the 3D city model generated images. Multiple image matching metrics were tested to find the pose of the generated image that is closest to the smartphone image chosen with great robustness.

Existing 3DMA vision-integrated approaches for urban positioning use either edge features or skyline for positioning. This study is a method that extends with both these paradigms to formulate the positioning as a semantic-based problem using material as the semantic information. Our experiments demonstrate that it is possible to outperform existing GNSS and advanced GNSS positioning methods in urban canyons. The advantages for the semantic-based VPS method are numerous:

- The formulation of positioning as a semantic-based problem enable us to apply the existing wide variety of advanced optimization/shape matching metrics to this problem.
- Materials is diverse, distinctive, and distributed everywhere, hence the semantic information in an image is easy to recognize.
- The utilization of building materials for positioning eliminates the need for skyline and building boundary reliance.
- Identification and consideration of foliage and dynamic objects such that it can be removed from positioning.
- Semantic of buildings stored as vector maps makes it simple to update and label accurately.

Considering the results presented in this paper, we conclude the proposed method improves upon the latitude, longitude and heading estimation of existing advanced positioning methods.

### 5.2. Future Work

Several potential future developments are suggested.

- Research has shown it is possible to identify a wide variety of materials in images in the indoor environment [39]. Therefore, it is suggested to develop and train a deep learning neural network to identify materials in smartphone images in the outdoor environment for practical use.

Improvement in the deep learning neural network could also aid automatic segmentation of 3D building models, reducing the offline preparation time.

- To add the six common building material classes to differentiate (including concrete, stone, glass, metal, wood, and bricks) means that given a large and high-quality dataset, the proposed method can be adapted to a variety of different uses.
- It is possible to provide computation of depth based on the 3D city model and the virtual camera, which can then be stored as additional information in the generated images. This information of depth can allow precise AR after image matching.
- To maximize all available visual information, the combination of semantic-based VPS and feature-based VPS could yield better positioning performance.
- To reduce storage and computational load, the images can be stored as contour coordinates instead of pixels.
- The semantic-based VPS could also be further improved by extending the functionality to work in different weather, time, and brightness conditions.
- One difficult encountered in this experiment is the discrepancy between reality and the 3D city model, hence it is suggested to use cloud-sourcing map to continuous update the model.
- For dynamic positioning, we can use a multiresolution framework, where the search starts from a big and sparse grid and is then successively refined on smaller and denser grids. Thus, the pose of the chosen candidate is used to refine a smaller search area.